\documentclass[a4paper,fleqn]{cas-dc}

\usepackage[authoryear,longnamesfirst]{natbib}

\def\tsc#1{\csdef{#1}{\textsc{\lowercase{#1}}\xspace}}
\tsc{WGM}
\tsc{QE}
\tsc{EP}
\tsc{PMS}
\tsc{BEC}
\tsc{DE}

\begin{document}
\let\WriteBookmarks\relax
\def\floatpagepagefraction{1}
\def\textpagefraction{.001}

\shorttitle{GDDS}


\title [mode = title]{GDDS: A Single Domain Generalized Defect Detection Frame of Open World Scenario using Gather and Distribute Domain-shift Suppression Network}                      

\tnotetext[1]{This work was partially supported by the National Key Research and Development Program of China under Grant 2022YFB3303800, and the National Natural Science Foundation of China under Grants U21A20482 and 62073117.}

%
\author[1]{Haiyong Chen}
\cormark[1]
\ead{haiyong.chen@hebut.edu.cn}
\affiliation[1]{organization={College of Artificial Intelligence and Data Science},
    addressline={Hebei University of Technology}, 
    city={Tianjin},
    postcode={300130}, 
    country={China}}

\author[1,3]{Yaxiu Zhang}
\ead{13453276296@163.com}

\author[1,3]{Yan Zhang}
\ead{718362133@qq.com}

\author[1]{Xin Zhang}
\ead{2023927@hebut.edu.cn}

\author[2,3]{Xingwei Yan}[style=chinese]
\cormark[1]
\ead{yanxingwei@nudt.edu.cn}
\affiliation[2]{organization={College of Electronic Science},
    addressline={National University of Defense Technology}, 
    city={Changsha, Hunan},
    postcode={410073}, 
    country={China}}
\affiliation[3]{organization={Tianjin Advanced Technology Research Institute},
    city={Tianjin},
    postcode={300459}, 
    country={China}}

\cortext[cor1]{Corresponding author}

\begin{abstract}
Efficient and intelligent surface defect detection of photovoltaic modules is crucial for improving the quality of photovoltaic modules and ensuring the reliable operation of large-scale infrastructure. However, the scenario characteristics of data distribution deviation make the construction of defect detection models for open world scenarios such as photovoltaic manufacturing and power plant inspections a challenge. Therefore, we propose the Gather and Distribute Domain shift Suppression Network (GDDS). It adopts a single domain generalized method that is completely independent of the test samples to address the problem of distribution shift. Using a one-stage network as the baseline network breaks through the limitations of traditional domain generalization methods that typically use two-stage networks. It not only balances detection accuracy and speed but also simplifies the model deployment and application process. The GDDS includes two modules: DeepSpine Module and Gather and Distribute Module. Specifically, the DeepSpine Module applies a wider range of contextual information and suppresses background style shift by acquiring and concatenating multi-scale features. The Gather and Distribute Module collects and distributes global information to achieve cross layer interactive learning of multi-scale channel features and suppress defect instance shift. Furthermore, the GDDS utilizes normalized Wasserstein distance for similarity measurement, reducing measurement errors caused by bounding box position deviations. We conducted a comprehensive evaluation of GDDS on the EL endogenous shift dataset and Photovoltaic inspection infrared image dataset. The experimental results showed that GDDS can adapt to defect detection in open world scenarios faster and better than other state-of-the-art methods.
\end{abstract}


\begin{highlights}
\item Analysis and proof of endogenous shift phenomenon in open world photovoltaic field scenarios.
\item A single domain generalized detection scheme that is completely independent of the test samples.
\item Efficient, high-speed, and adaptable solutions can improve the detection performance and universal performance of multi-scale and multi-target detection.
\item Analysis and Improvement of Boundary Box Deviation
\end{highlights}

\begin{keywords}
Single-domain generalized object detection \sep open world scenario \sep domain shift \sep endogenous shift \sep Photovoltaic manufacturing \sep Unmanned aerial vehicle inspection
\end{keywords}
\maketitle

\section{Introduction}
\label{sec:introduction}

With the continuous growth in global demand for renewable energy, photovoltaics (PV), as efficient solar energy conversion devices, play an irreplaceable role in the global energy structure transition. However, various defects unavoidably occur in PV modules during production, transportation, installation, and operation processes \cite{b1}, \cite{b2}. The induced potential decay can significantly reduce the photovoltaic conversion efficiency of PV modules (greater than 30\%) \cite{b3}, \cite{b4}. Over the long term, severe defects may lead to hot spot effects, increase the risk of fires in solar panels, and result in significant property damage \cite{b5}. Therefore, to ensure the performance and safety of photovoltaic panels and guarantee the reliable operation of large-scale photovoltaic power plants, efficient and intelligent detection of surface defects in photovoltaic modules is crucial.

\begin{figure}
	\centering
		\includegraphics[scale=0.33]{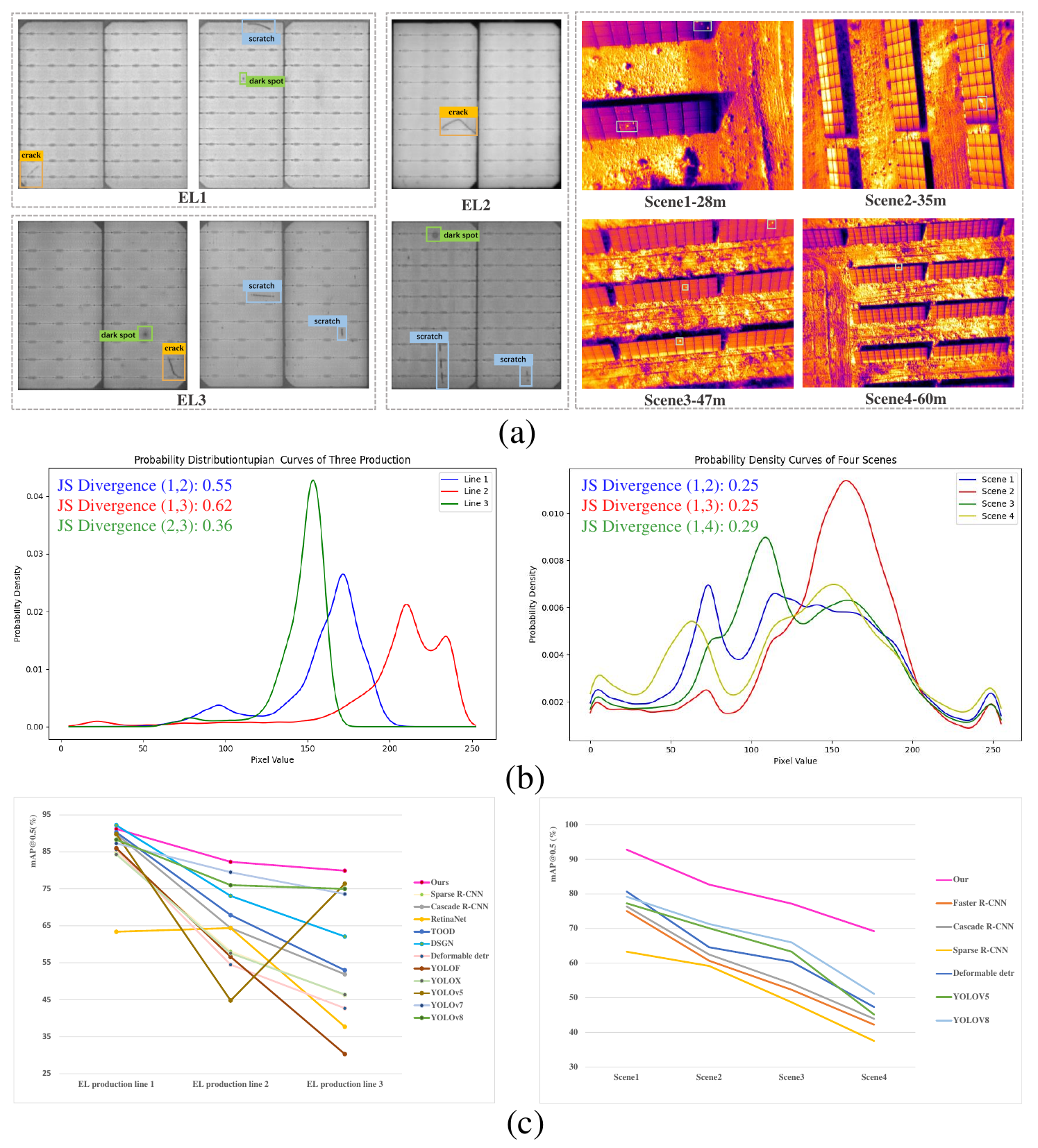}
	\caption{Endogenous shift phenomenon in open world scenarios. (a) In open world scenarios, examples of images acquired from different domains are shown. The left image displays EL images obtained from various production lines in photovoltaic manufacturing scenes. Defective images obtained from different production lines exhibit slight differences in texture, clarity, brightness, and other aspects. The right image shows infrared images acquired at different heights in photovoltaic inspection scenarios. (b) The left and right images respectively depict probability density distribution curves of data from different domains in photovoltaic manufacturing and photovoltaic inspection scenarios. This illustrates the differences in data distribution among different domains. (c) Different detectors were trained on the source domain and tested on multiple other domains. The left and right figures respectively represent scenarios of photovoltaic manufacturing and photovoltaic plant inspection.}
	\label{Fig.1}
\end{figure}

However, in the practical application of surface defect detection in photovoltaic modules, the phenomenon of endogenous shift \cite{b6} occurs due to the influence of various factors. Including but not limited to natural factors such as weather changes, wind and rain, dust, material aging and degradation, imaging mechanisms, inspection heights, etc. As shown in Fig. \ref{Fig.1}, significant differences exist in inspection images obtained at different heights during photovoltaic plant inspections. This phenomenon of distribution shift significantly reduces the stability and robustness of defect detection algorithms, resulting in poor performance of the algorithms in actual industrial production environments. In the photovoltaic manufacturing process, images of the same type of photovoltaic components collected from different production batches exhibit relatively minor endogenous shift as perceived by the naked eye. However, when measuring the similarity of data from different production lines using JS divergence, it is found that the degree of endogenous shift is even more severe than the distribution shift of inspection images from different scenes in photovoltaic plants. This is because this scene is significantly affected by environmental changes, production line changes, equipment aging, hardware components, imaging mechanisms, operational errors, process innovations, raw material differences, temperature, pressure, and lighting factors. The endogenous shift caused by this phenomenon leads to missed or false detections, hindering effective management and handling of surface defects in photovoltaic components, thus affecting the long-term stable operation of the system.

Due to the differentiated characteristics of data distribution in such scenarios, building defect detection models for open world scenario poses a challenge. Existing traditional models \cite{b7}--\cite{b9} often fail to handle changes in data distribution and cannot fully adapt to the complexity and diversity of such scenes. Therefore, more intelligent and flexible approaches are needed to address this issue. Domain adaptation (DA) \cite{b10}, multi-source domain generalization (MDG) \cite{b11}, and many other methods have been studied to learn domain-invariant features. However, these methods often require a large amount of domain-specific knowledge and manual feature engineering, with their performance heavily relying on the expertise of domain experts, limiting their generalizability across different environments and applications. Additionally, in practical applications, it is often difficult or even unknown to acquire data before deploying the model.

In contrast, single-domain generalization \cite{b12} utilizes only data from a single source domain to train the model, enhancing its ability to generalize to other unseen domains. This approach offers a more flexible and adaptive solution, although it is rarely adopted to address the distribution shift problem \cite{b13} in open world scenario such as photovoltaic manufacturing and station inspections. Therefore, this paper proposes a novel single-domain general defect detection framework named GDDS. The aim is to effectively address the distribution shift problem in open world scenario such as photovoltaic manufacturing and station inspections. Additionally, traditional domain generalization methods typically employ a two-stage network as the baseline network, separately conducting feature learning and transfer learning. Two-stage network models require more data and computational resources, have high complexity, and are challenging to tune. Moreover, the two-stage process can lead to information loss or redundancy, impacting the model's performance. Therefore, we propose using a single-stage network as the baseline network, achieving feature extraction and classification in an end-to-end manner. This approach rapidly adapts to changes in data distribution in open world scenario, making it more suitable for large-scale industrial applications.The main contributions of this paper are summarized as follows:

\begin{enumerate}
\itemsep=0pt
\item We investigated and demonstrated the phenomenon of endogenous shift in open world scenario such as photovoltaic manufacturing and station inspections. We proposed to address this issue using a single-domain generalization approach that is entirely independent of the target domain. For the first time, we employed a one-stage network that balances accuracy and speed as the baseline network, replacing the two-stage networks typically used in traditional domain generalization methods. To the best of our knowledge, there are few methods currently adopted in the industry to address the phenomenon of endogenous drift in this way.
\item To enhance the robustness of defect target detection in open dynamic scenes, we introduced GDDS. GDDS consists of two modules: the DeepSpine Module (D2) and the Gather and Distribute Module (GD), which suppress background drift and instance drift, respectively. In bounding box regression, we utilized normalized Wasserstein distance for similarity measurement to reduce measurement errors caused by bounding box position deviations.
\item We comprehensively evaluated GDDS on the EL endogenous shift dataset and Photovoltaic inspection infrared image dataset. The results indicate that GDDS outperforms other state-of-the-art methods.
\end{enumerate}

The remaining sections of the paper are organized as follows: section II specifically enumerates the relevant work on photovoltaic cell defect detection both domestically and internationally. In response to the endogenous shift issue in photovoltaic cell defect detection, our proposed GDDS is detailed in section III. Some experimental details of executing this method are showcased in section IV. The experimental results and analysis of the GDDS are presented in section V. Lastly, section VI provides the conclusion.

\section{RELATED WORK}
\label{sec:RELATED WORK}

\subsection{Deep Learning Approaches for Photovoltaic Defect Detection}

In recent years, deep convolutional neural networks (CNNs) have shown remarkable capabilities in self-learning, fault-tolerance, and adaptability. They have been successfully applied to detect defects in photovoltaic (PV) cell images. Li et al. \cite{b14}--\cite{b15} used CNNs to extract deep features from operational PV modules, improving the efficiency and accuracy of defect classification. Deitscha et al. \cite{b7} employed end-to-end deep CNNs to detect defects in EL images, achieving higher accuracy than manual methods, enabling continuous monitoring of PV cells. Akram et al. \cite{b16} proposed a light CNN method for defect identification in EL images, optimizing for computational efficiency without compromising real-time performance. However, supervised learning with limited data may lack stability and generalization. Akram addressed this by using data augmentation, but these methods are not robust in complex environments. Chen et al. \cite{b17} designed a multispectral deep CNN for defect detection, enhancing accuracy and adaptability. Su et al. \cite{b8} introduced a bidirectional attention network and an improved detector, along with a residual channel attention gate network (RCAG-Net) \cite{b9} to suppress backgrounds and highlight defects. However, these methods overlook the endogenous shift problem in real PV manufacturing scenarios, resulting in performance degradation and unsatisfactory results in practical settings.

\subsection{Single-Domain Generalization for Photovoltaic Defect Detection}
In real-world scenarios with domain shift, various methods aim to improve model generalization by learning domain-invariant features. For instance, Ding et al. \cite{b18} proposed a transfer learning-based solution that extracts defect features across different levels of representation, exhibiting robustness. However, these methods' performance depends on training sample size, with larger samples improving performance but fewer samples leading to overfitting. Acquiring multi-domain training data is challenging and costly, making it impractical to obtain data from the latest domains in advance. Zhao et al. \cite{b13} utilized a single-domain generalization approach aligning deep and shallow style features to suppress background shift and employing graph convolution to capture richer semantic dependencies, achieving excellent generalization effects. However, their two-stage object detection method was computationally complex and slow. To efficiently operate photovoltaic panels and meet high-speed, high-precision requirements for defect detection, we propose a novel single-stage network-based defect detection model aiming to learn richer global information representations to suppress endogenous shift on different production lines.

\section{METHODOLOGY}

\begin{figure}
	\centering
		\includegraphics[scale=0.4]{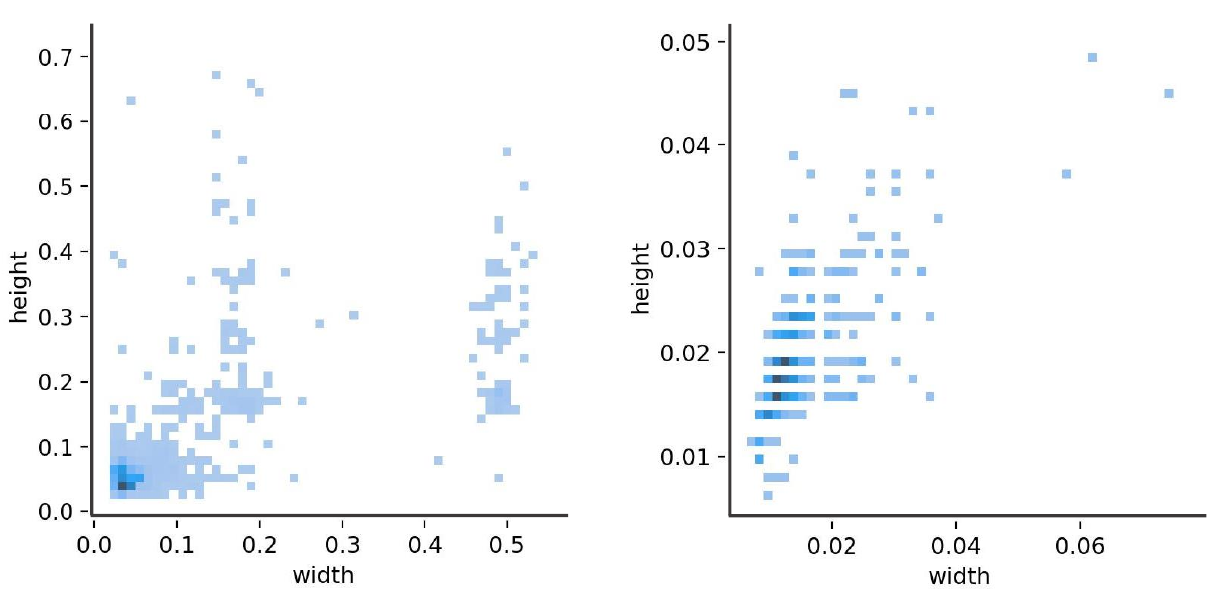}
	\caption{The left and right figures respectively depict the distribution of defect sizes in photovoltaic manufacturing and photovoltaic plant inspection scenarios. The horizontal axis represents the ratio of the actual width of target defect bounding boxes to the width of EL images, while the vertical axis represents the ratio of the actual height of target defect bounding boxes to the height of EL images.}
	\label{Fig.2}
\end{figure}

\subsection{DeepSpine Module}

The diversity and complexity of photovoltaic defects result in a wide range of scales. As shown in Fig. \ref{Fig.2}, there are predominantly small defects \cite{b19}, but also some medium and large defects. The broad scale range of photovoltaic defects poses a significant challenge for defect detection in photovoltaic cell manufacturing. Due to factors such as network layers, the original one-stage network is more suited to detecting medium-sized defects, and its performance is poorer for small defects, accounting for less than 20\% of the image, and large defects, accounting for more than 60\% of the image. Therefore, it is proposed to increase the levels of the backbone network and concatenate shallower feature maps with deeper feature maps. On the one hand, a deeper backbone network enables the model to capture and integrate information across multiple scales, facilitating a better response to defects of various sizes and enhancing generalization. On the other hand, lower-resolution feature maps can capture a broader context, and by using a small object detection layer to detect features concatenated from shallower and deeper feature maps, the model can fully utilize global contextual information. This helps the model better distinguish between targets and backgrounds in complex backgrounds, adaptively suppress background shifts in any domain, and improve adaptability to complex environments.

\begin{figure*}
	\centering
		\includegraphics[scale=0.46]{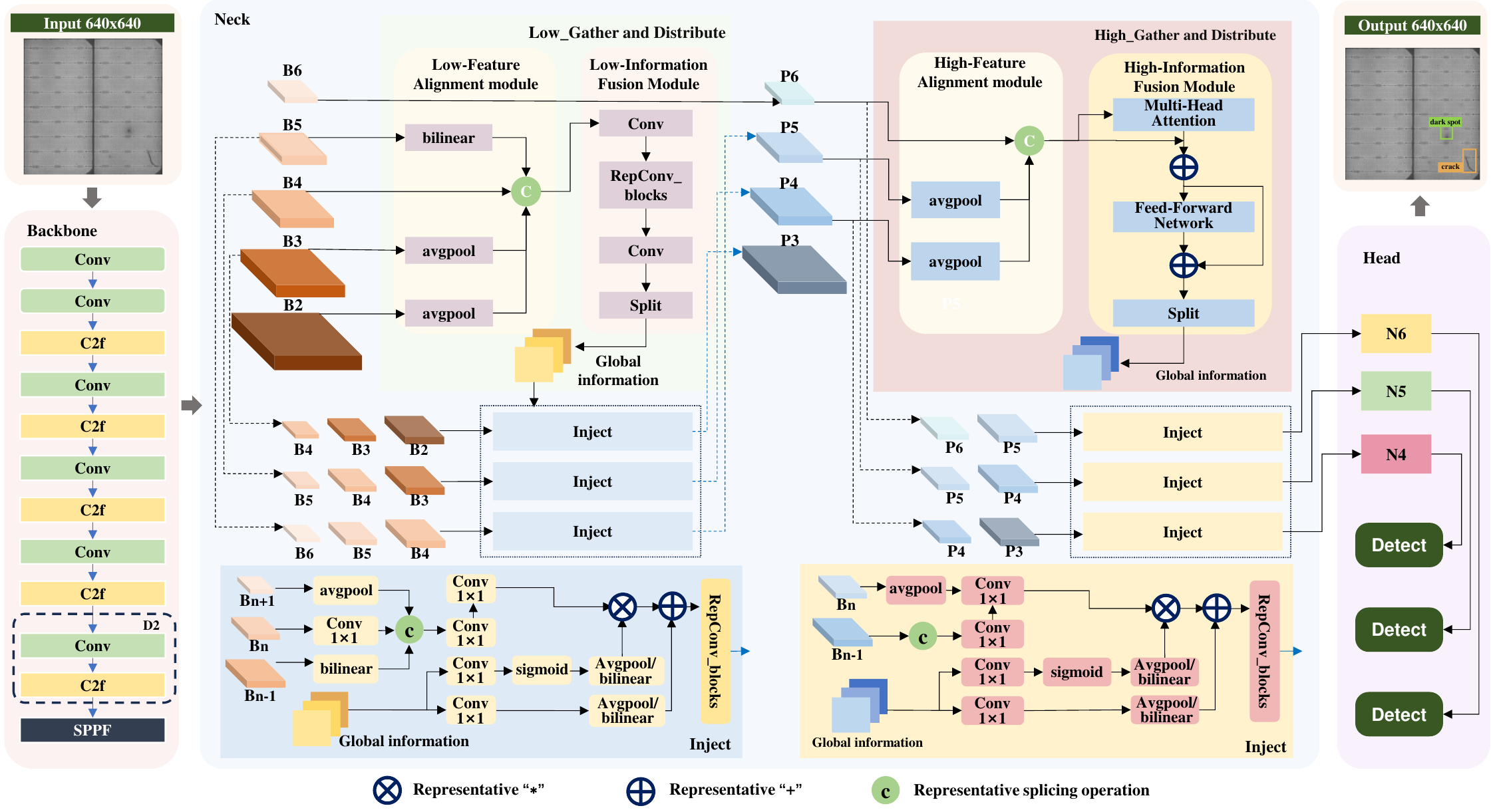}
	\caption{Overall network architecture.}
	\label{Fig.3}
\end{figure*}

\subsection{Gather and Distribution Module}

Learning rich multi-level information and semantic dependencies is crucial for enhancing a model's ability to localize and discriminate in unknown domains. Contextual information across different scales is complementary, and fully leveraging contextual and global information enables the acquisition of semantic dependencies among multi-scale channel feature maps. However, the original Feature Pyramid Network structure suffers from information loss due to its hierarchical information fusion pattern, where adjacent layers undergo sufficient fusion, and cross-layer fusion relies only on intermediate layers through recursive fusion. The indirect acquisition of information leads to some information loss. Therefore, as shown in Fig. \ref{Fig.3}, this paper proposes an information aggregation-distribution mechanism for multi-scale information interaction and fusion. Different levels of features are globally fused to obtain global information. Then, inject global information into features at different levels to establish a direct path for cross layer information fusion. And increase the full integration of deeper and shallower information, improving the generalization performance for both maximum and minimum targets. Realize efficient information exchange and fusion between multi-scale channels. This enhances the spatial contextual perception capability of defect targets and strengthens the spatial representation of defect targets. It helps in better understanding the position, shape, and relative relationships of targets in the image, obtaining multi-level feature information including defect feature information, spatial information, etc., for distinguishing defects. Adaptive learning of rich discriminative defect feature information improves the discriminative and positioning capabilities when facing defects in unknown domains, effectively suppressing instance deviation of defects, and enhancing the generalization capability of the model.

\begin{figure}
	\centering
		\includegraphics[scale=0.3]{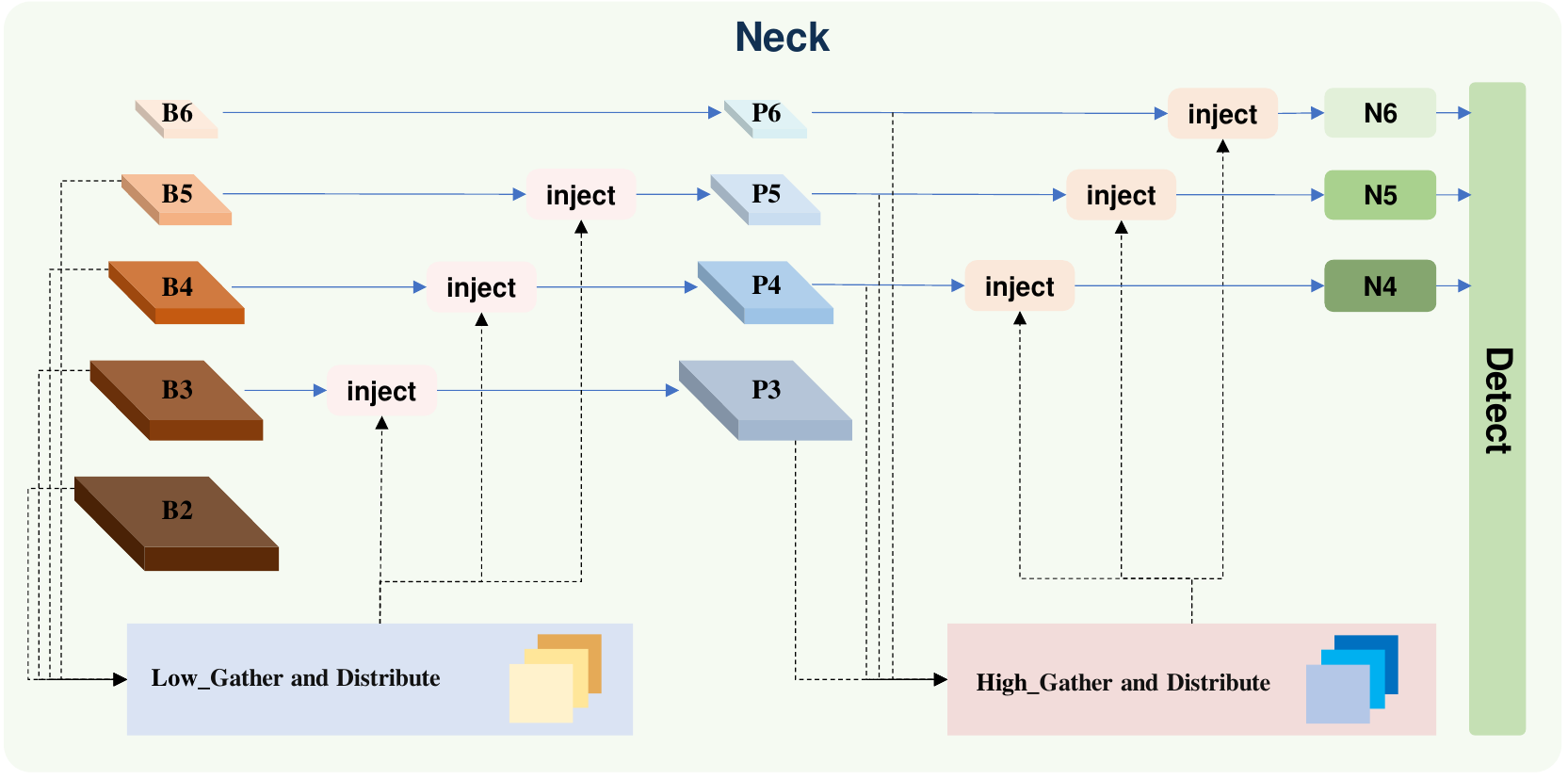}
	\caption{Overall network architecture.}
	\label{Fig.4}
\end{figure}

\subsection{Gather and Distribution Module}

At the same time, as shown in Fig. \ref{Fig.2}, in order to enhance the model's ability to detect multi-scale defective targets, low-level gathering and distribution mechanism (Low-GD) and high-level gathering and distribution mechanism (High-GD) are used to respectively extract and fuse feature maps of large and small sizes. Especially by increasing the fusion of the p2 and p6 layers, it fully utilizes the information from shallower and deeper layers. Improved the accuracy and robustness of the network on both maximum and minimum targets. Furthermore, the information aggregation-distribution mechanism's information interaction fusion mechanism consists of three modules\cite{b20}.

Firstly, the Feature Alignment module collects feature information from different levels. It achieves alignment of feature information from different levels by bilinear interpolation on deeper features and average pooling on shallower features, unifying their sizes.

Secondly, the Information Fusion Module integrates the feature information unified by the Feature Alignment module and distributes the fused global information into local information. The low-level information fusion module utilizes Reparameterized Convolution Blocks (RepConvblocks) to transform the number of channels of global features into the sum of channels of local features to be injected. This allows segmented global features to adapt to different model sizes and further fuse with features at different levels. The high-level information fusion module employs a multi-head attention mechanism, Feedforward Network, and concatenation operation to achieve the fusion of deep-level information.

Finally, the Information Injection Module injects the fused feature information into different levels after the fusion process in the Information Fusion Module is completed. It utilizes attention to merge local and global features. When the sizes of local and global features differ. It still employs average pooling for down-sampling or bilinear interpolation for up-sampling to resize the features, ensuring proper alignment. At the end of each attention fusion, RepConvblocks are used to further extract and fuse information.

\subsection{Loss Function}

\begin{figure*}
	\centering
		\includegraphics[scale=0.33]{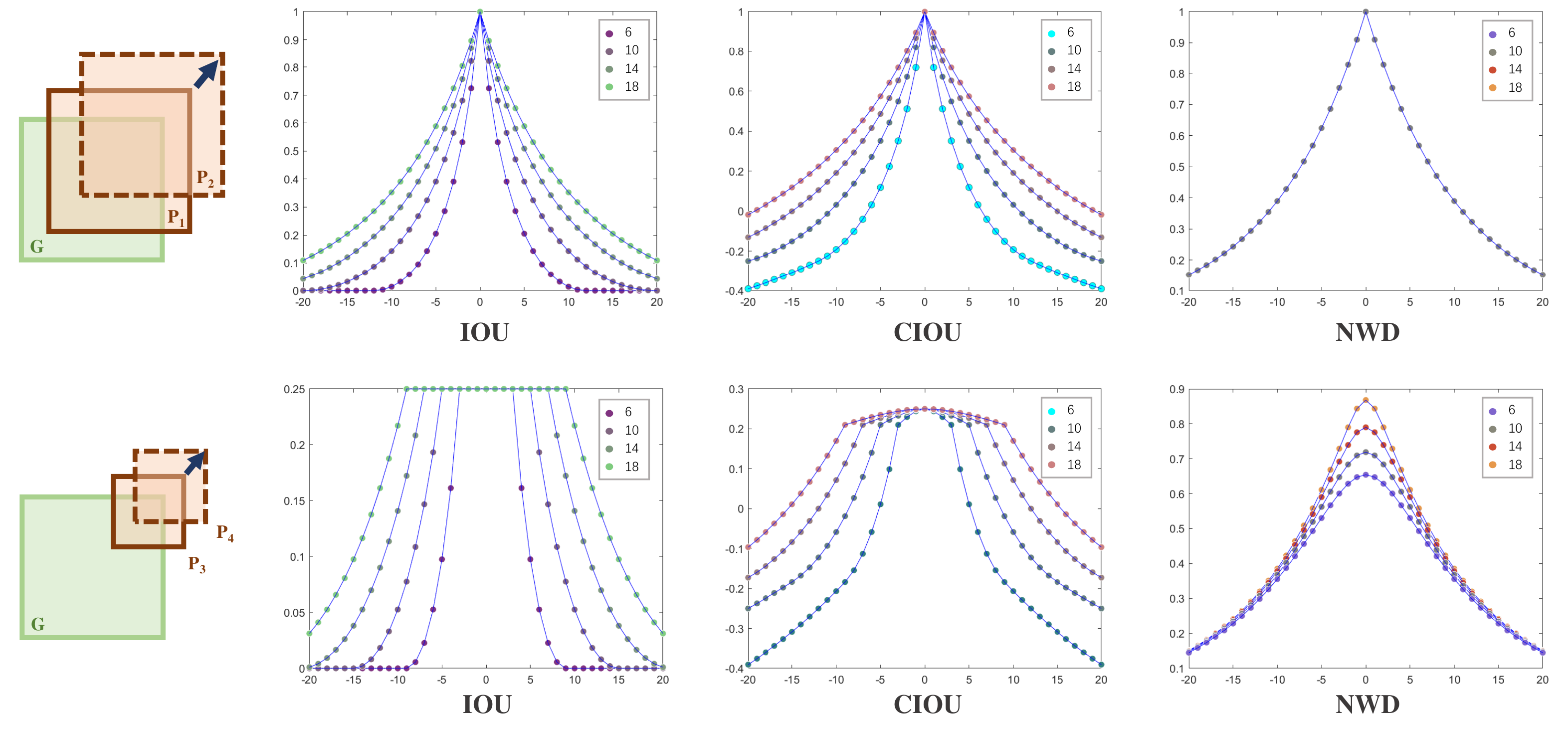}
	\caption{The impact of bounding box deviation on different loss calculation results. $ G $ represents the true bounding box, and $ P $ represents the predicted bounding box. Points of different colors represent defect targets of different sizes. The horizontal axis values represent the pixel difference between the predicted bounding box and the true bounding box center points, while the vertical axis values represent the corresponding measurement values.}
	\label{Fig.5}
\end{figure*}

Bounding box regression is a crucial step in object detection. However, as shown in Fig. \ref{Fig.5}, the positional deviation of the predicted bounding boxes for minor defects in photovoltaic components is still significant with CIOU, and the CIOU values between small predicted bounding boxes and true bounding boxes are lower than the minimum positive threshold. The average number of positive samples assigned to each true bounding box is less than 1, which is not a good measure for the existence of minor defects in photovoltaic components. Therefore, we adopt an improved method for calculating the loss function, using the Wasserstein distance(NWD) \cite{b21} to measure the similarity of bounding boxes instead of CIOU, reducing the sensitivity to positional deviations for small-scale defect targets.

First, the predicted bounding box $ P = \left(b x^{p}, b y^{p}, w^{p}, h^{p}\right)$ and the true bounding box $ G = \left(b x^{gt}, b y^{gt}, w^{gt}, h^{gt}\right)$ are modeled as two-dimensional Gaussian distributions, obtaining $ {~N}^P $ and $ {~N}^G $. The second-order Wasserstein distance between $ {~N}^P $ and $ {~N}^G $ can be expressed as

\begin{align}
\label{eq1}
\begin{aligned}
&W_2^2\left(~N^P, ~N^G\right) \\
&=\left\|\left(\left[b x^{p}, b y^{p}, \frac{w^{p}}{2}, \frac{h^{p}}{2}\right]^T,\left[b x^{gt}, b y^{gt}, \frac{w^{gt}}{2}, \frac{h^{gt}}{2}\right]^T\right)\right\|_2^2.
\end{aligned}
\end{align}

Using the Wasserstein distance to measure the distance between the predicted bounding box and the true bounding box cannot be directly applied as a similarity measure. Therefore, it is normalized to obtain the normalized Wasserstein distance, which can measure the similarity of bounding boxes.

\begin{align}
\label{sjd}
NWD\left({N}^P, {~N}^G\right)=\exp \left(-\frac{\sqrt{W_2^2\left({~N}^P, {~N}^G\right)}}{C}\right),
\end{align}

where $C$ is a constant, typically set to the average absolute size of targets in the dataset to achieve optimal performance. According to our datasets, we set $C = 43$ in our experiments. The complete definition of the NWD loss function is:

\begin{align}
L_{N W D}=1-N W D\left({~N}^P, {~N}^G\right).
\end{align}

The NWD not only considers the spatial structure between bounding boxes but also takes into account the consistency of feature distributions. Even in cases where there is no overlap or the overlap is negligible, it can still measure the similarity of distributions. Specifically, as depicted in the first row of Figure \ref{Fig.5}, assuming the predicted box is the same size as the ground truth box, the deviation of the predicted bounding box from the center point along the diagonal direction of the ground truth bounding box is shown. Experimental results indicate that IOU and CIOU calculations are highly sensitive to position deviations compared to NWD. The four curves based on NWD completely overlap and exhibit smoother relative changes. In the second row of the figure, assuming the predicted box size is half the size of the ground truth box, NWD calculations still result in much smoother curves compared to IOU and CIOU calculations under the same offset conditions. This demonstrates that NWD is insensitive to multi-scale defect targets, exhibiting a certain degree of smoothness and adaptability to position deviations.

In this study, we use a combination of CIOU and NWD as the final measurement method, with the calculation as follows:

\begin{align}
L=L_{\mathrm{ClOU}}+\beta L_{N W \mathrm{D}}
\end{align}

In the experimental section, we evaluated the impact of the balancing factor value $\beta$ on the detection performance. Through experimentation, we observed that the detection performance is optimized when $\beta$ is set to 0.5.

\section{EXPERIMENTAL RESULTS}

In this section, we conducted a large number of experiments to evaluate GDDS. In order to assess the effectiveness of this method in suppressing endogenous shift in dynamic open scenarios, we conducted detection experiments on two datasets with different degrees of distribution shift: the EL Endogenous Shift Dataset and the Photovoltaic Inspection Infrared Image Dataset.

\subsection{EL Endogenous Shift Dataset}

\begin{figure}
	\centering
		\includegraphics[scale=0.3]{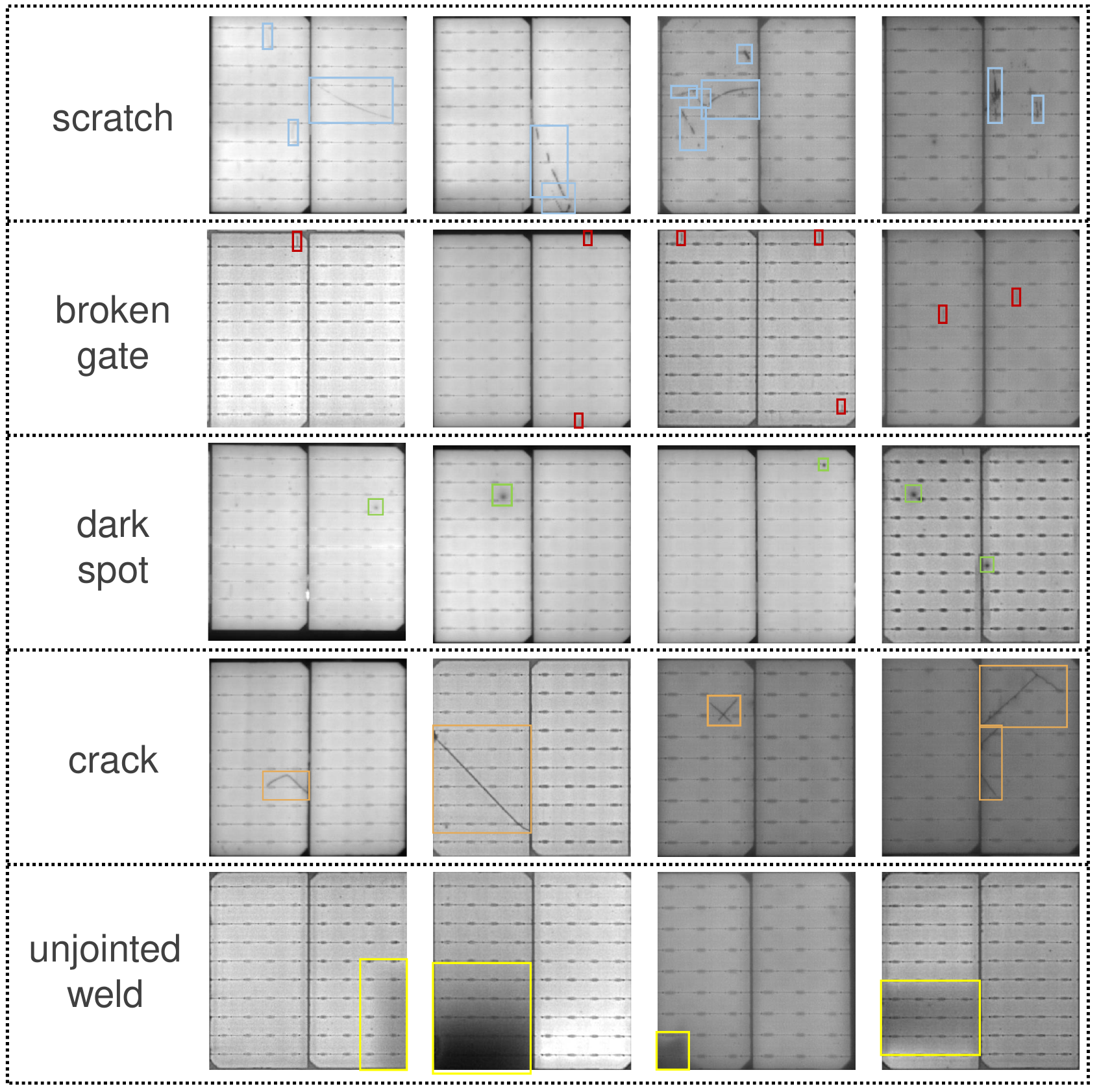}
	\caption{Examples of different types of defects. The first column in the image represents the names of the defects. Different colored bounding boxes represent the true bounding boxes of defects belonging to different categories.}
	\label{Fig.6}
\end{figure}

\begin{figure}
	\centering
		\includegraphics[scale=0.2]{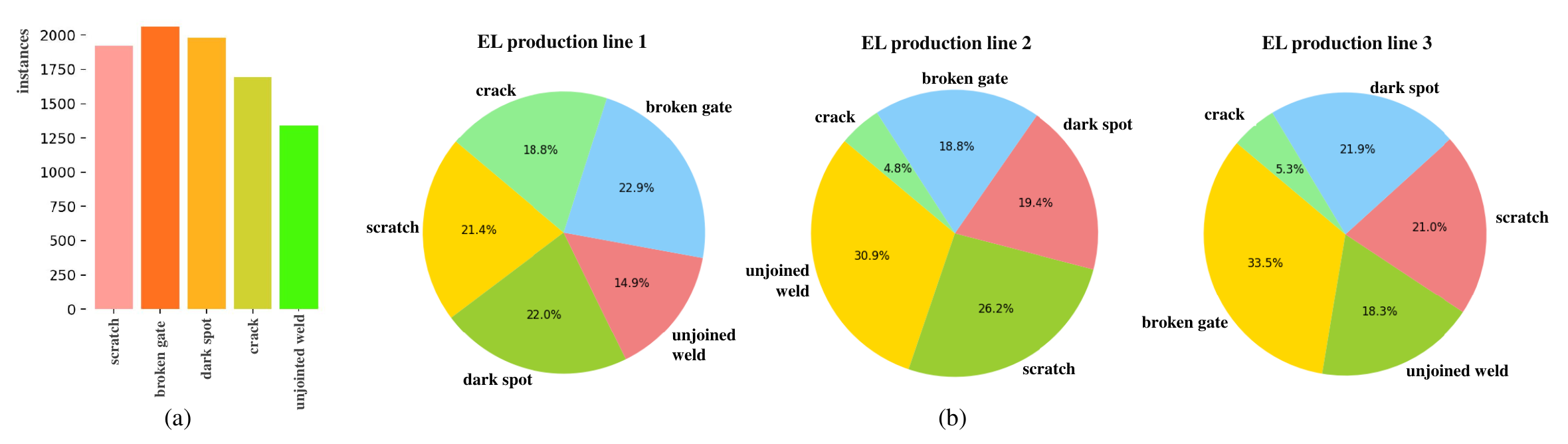}
	\caption{(a) Summary table of the total number of various types of defects. (b) Schematic diagram of the distribution of various types of defects. The proportion of defect categories varies between different production lines.}
	\label{Fig.7}
\end{figure}

This study utilized electronic luminescence images of PV module components from three production lines as an endogenous shift dataset. A total of 16,323 images were included. The allocation of training and testing sets is shown in Table \ref{tabbe l}. The dataset comprises five types of defects: soldering open, scratching, hidden cracks, dark spots, and broken grid. We annotated the real bounding boxes of defects using the labeling tool "labelimg". The samples and distribution of various types of defects are illustrated in Fig. \ref{Fig.6} and Fig. \ref{Fig.7}.

\subsection{Photovoltaic inspection infrared image dataset}

\begin{figure}
	\centering
		\includegraphics[scale=0.3]{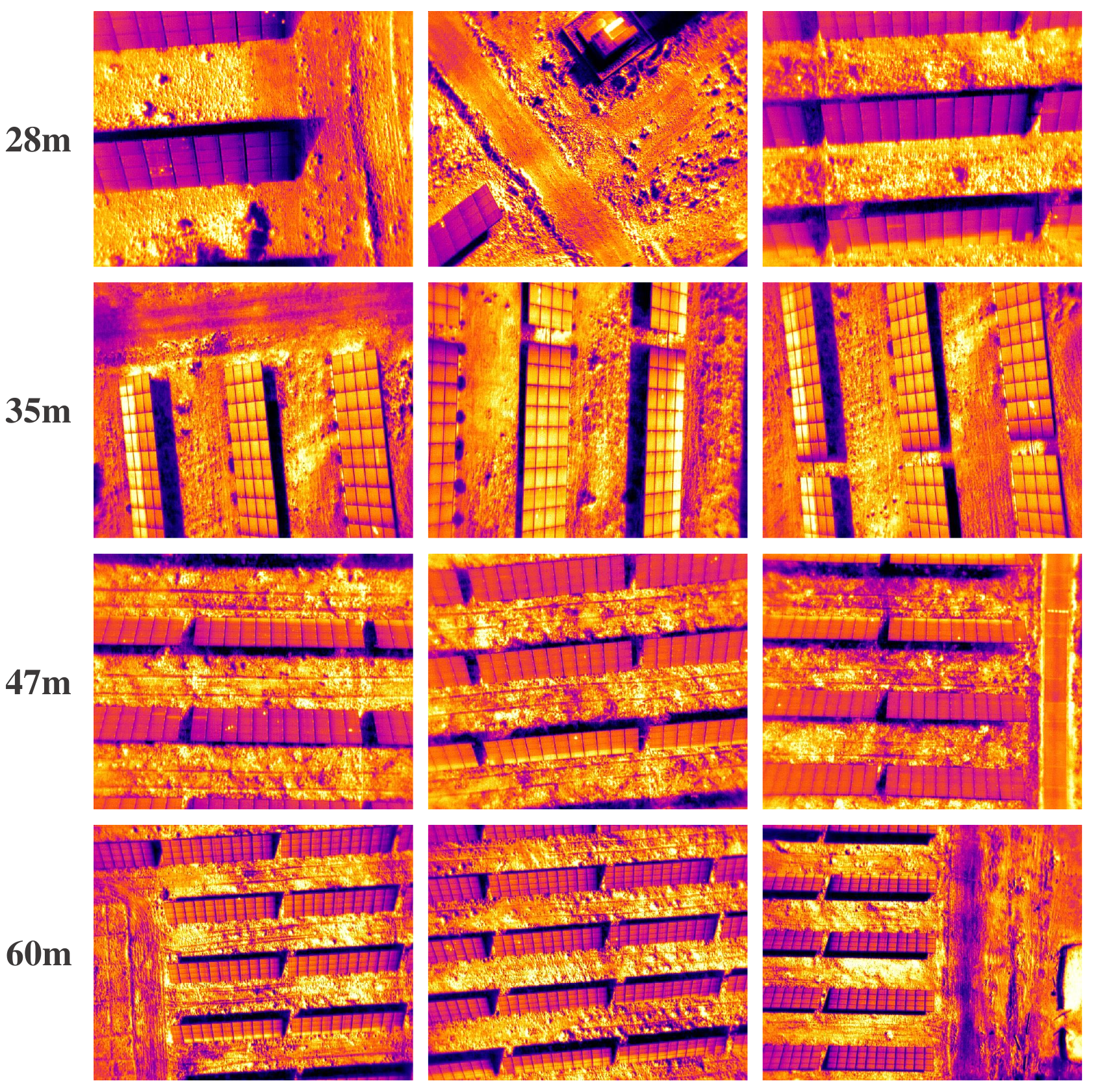}
	\caption{Examples of infrared images for photovoltaic detection at four different heights. Images captured at different heights contain varying numbers of photovoltaic components. Additionally, the proportion of individual photovoltaic components to the image size also varies accordingly.}
	\label{Fig.8}
\end{figure}

\begin{table}[htbp]\scriptsize
  \centering
  \caption{Allocation of EL Endogenous Shift Dataset}
    \begin{tabular}{cccccc}
    \hline
    Datasets & Total & Training & Testing & Resolution & Acquisition time \\
    \hline
    EL 1  & 9172  & 6652  & 2520  & 384×384 & September 2020 \\
    EL 2  & 5750  & \textbackslash{} & 5750  & 640×589 & May 2021 \\
    EL 3  & 1401  & \textbackslash{} & 1401  & 700×668 & October 2021 \\
    \hline
    \end{tabular}%
  \label{tabbe l}%
\end{table}%

As shown in Fig. \ref{Fig.8}, this dataset consists of infrared images of PV panels acquired using infrared thermography by drones at four heights: 28m, 35m, 47m, and 60m. There are a total of 7552 images, each sized 640×512 pixels. Specifically, in the scenario captured at a height of 28 meters, 2152 images were taken, with 1365 images used for training and the remaining 787 images serving as the test set for scenario 1. In the scenario captured at 35 meters high, 2115 images were taken and used as the test set for scenario 2. In the scenario captured at 47 meters high, 1674 images were taken and used as the test set for scenario 3. In the scenario captured at 60 meters high, 1611 images were taken and used as the test set for scenario 4. Furthermore, this dataset contains only one category of defects.

\subsection{Evaluation Metrics}

To assess the generalization performance of the model on the endogenous shift dataset, precision (P), recall (R), and mean Average Precision (mAP@0.5) at an intersection over union threshold of 0.5 between true bounding boxes and predicted bounding boxes are utilized. Precision represents the proportion of correctly detected targets, while recall represents the proportion of correctly predicted targets among all targets. The calculation methods are as follows:

\begin{align}
P=\frac{\mathrm{TP}}{\mathrm{TP}+\mathrm{FP}},
\end{align}
\begin{align}
P=\frac{\mathrm{TP}}{\mathrm{TP}+\mathrm{FN}},
\end{align}
\begin{align}
\mathrm{mAP}=\frac{1}{S} \sum_{i=1}^S \mathrm{AP}_i,
\end{align}

where $TP$ represents the number of true positive samples predicted as positive. FP represents the number of true negative samples predicted as positive. FN represents the number of false negative samples predicted as negative. S represents the total number of target classes.

Additionally, we employ the parameters quantity (Parameters) and the computational complexity (Giga Floating-point Operations Per Second, GFLOPS) to evaluate the model's speed. GFLOPS refers to the number of floating-point operations per second, measured in billions.

\subsection{Implementation Details}

We trained the network on an NVIDIA GeForce RTX 2060S GPU with an 11th Gen Intel(R) Core (TM) i7-11700K @ 3.60GHz 3.60 GHz processor. Python 3.8 programming language was used, and the deep learning network was constructed using the PyTorch 1.7.0 framework. The input image size was adjusted to 640 × 640. The model was trained for 150 epochs, with 16 preprocessed images input per iteration. Stochastic gradient descent (SGD) was chosen as the optimizer, with an initial learning rate set to 0.01, momentum set to 0.937, and a decay factor of 0.0005. We train the model exclusively on the training set and evaluate it on the remaining production lines or other high-altitude test sets. Since most samples in real-world scenarios are defect-free, we reserve defect-free samples for testing. Additionally, during training, we utilize various image augmentation techniques to enhance the data, including vertical flipping, horizontal flipping, mirroring, and others.

\subsection{Comparison With Recent Works}

To validate the superior performance of the proposed model, we compared it with the state-of-the-art models on the endogenous shift dataset. The compared models encompass a wide range of high-performance two-stage networks, one-stage networks, and Transformer-based methods. Specifically, we included Sparse R-CNN, Cascade R-CNN, DGGD, YOLOF, YOLOX, YOLOV5, YOLOV8, RetinaNet, TOOD, and Deformable DETR. Table \ref{table 2}, Table \ref{table 3} and Table \ref{table 4} present the experimental results on production lines 1, 2, and 3, respectively.

\renewcommand{\arraystretch}{1.4} 
\setlength{\tabcolsep}{3pt} 

\begin{table}[htbp]
  \centering
  \scriptsize 
  \caption{Comparison of experimental performance of different methods on EL production line 2}
  \label{table 2}
  \begin{tabular}{m{3em}<{\centering} m{5em}<{\centering} m{2.5em}<{\centering} m{2.5em}<{\centering} m{2.5em}<{\centering} m{2.5em}<{\centering} m{2.5em}<{\centering} m{2.5em}<{\centering} m{2.5em}<{\centering}}
    \hline
    \multicolumn{2}{c}{Detector} & scratch & \makecell[c]{broken\\gate} & \makecell[c]{dark\\sport} & crack & \makecell[c]{unjointed\\weld} & all & GFLOPS \\
    \hline
    \multirow{3}[2]{*}{\makecell[c]{Two-stage\\detectors}} & Sparse R-CNN & 87.3 & 91.1 & 84.3 & 74.1 & 97.5 & 84.9 & 61.9 \\
          & Cascade R-CNN & 92.5 & 95.9 & 89.6 & 78.7 & 92.4 & 89.8 & 55.5 \\
          & SSN & 94.2 & 96.6 & 91 & 83.9 & 95.3 & 92.2 & 60.4 \\
    \hline
    \makecell[c]{Trans-\\former\\method} & Deformable detr & 89.1 & 86.1 & 90.8 & 75.2 & 86.6 & 85.6 & 65.7 \\
    \hline
    \multirow{8}[4]{*}{\makecell[c]{One-stage\\detectors}} & RetinaNet & 78.5 & 77.4 & 34.4 & 50.4 & 76.1 & 63.4 & 38.2 \\
          & TOOD & 93.2 & 95.7 & 91.5 & 76.5 & 94.4 & 90.3 & 56.8 \\
          & YOLOF & 87.5 & 93.7 & 89.6 & 65.8 & 93.7 & 86 & 15.1 \\
          & YOLOX & 90.4 & 87.7 & 81.7 & 71.2 & 90.5 & 84.3 & 14.6 \\
          & YOLOV5 & 81 & 95.2 & 92.1 & 94.8 & 85.9 & 89.8 & 7.1 \\
          & YOLOV7 & 78.4 & 93 & 86.3 & 93.4 & 85.3 & 87.3 & 10.3 \\
          & YOLOV8 & 82.4 & 94.4 & 88.1 & 92 & 84.8 & 88.3 & 8.2 \\
\cline{2-9}          & \textbf{Ours} & \textbf{84.3} & \textbf{95.2} & \textbf{91.8} & \textbf{94.4} & \textbf{90.2} & \textbf{91.2} & \textbf{4.9} \\
    \hline
  \end{tabular}
\end{table}

\begin{table}[htbp]
  \centering
  \scriptsize 
  \caption{Comparison of experimental performance of different methods on EL production line 2}
  \label{table 3}
  \begin{tabular}{m{3em}<{\centering} m{5em}<{\centering} m{2.5em}<{\centering} m{2.5em}<{\centering} m{2.5em}<{\centering} m{2.5em}<{\centering} m{2.5em}<{\centering} m{2.5em}<{\centering} m{2.5em}<{\centering}}
    \hline
    \multicolumn{2}{c}{Detector} & scratch & \makecell[c]{broken\\gate} & \makecell[c]{dark\\sport} & crack & \makecell[c]{unjointed\\weld} & all & GFLOPS \\
    \hline
    \multirow{3}[2]{*}{\makecell[c]{Two-stage\\detectors}} & Sparse R-CNN & 64 & 69.5 & 58.3 & 63.2 & 35.3 & 58.1 & 61.9 \\
          & Cascade R-CNN & 67.3 & 69.7 & 67.5 & 68.1 & 49.3 & 64.4 & 55.5 \\
          & SSN & 71 & 74.9 & 85.3 & 75.8 & 58.2 & 73.1 & 60.4 \\
    \hline
    \makecell[c]{Trans-\\former\\method} & Deformable detr & 62.9 & 57.7 & 70.6 & 53.9 & 27.5 & 54.5 & 65.7 \\
    \hline
    \multirow{8}[4]{*}{\makecell[c]{One-stage\\detectors}} & RetinaNet & 67.3 & 69.7 & 67.5 & 68.1 & 49.3 & 64.4 & 38.2 \\
          & TOOD & 72.2 & 69 & 81.9 & 65.7 & 50.5 & 67.9 & 56.8 \\
          & YOLOF & 66.8 & 70.2 & 80.8 & 46.4 & 18.8 & 56.6 & 15.1 \\
          & YOLOX & 60.1 & 70.2 & 81.8 & 58.5 & 17.5 & 57.6 & 14.6 \\
          & YOLOV5 & 35.9 & 46.4 & 43.8 & 31.4 & 66.4 & 44.8 & 7.1 \\
          & YOLOV7 & 74.8 & 87.4 & 87.8 & 63.3 & 84.4 & 79.5 & 10.3 \\
          & YOLOV8 & 72.8 & 88.3 & 93.2 & 41.3 & 84.7 & 76 & 8.2 \\
\cline{2-9}          & \textbf{Ours} & \textbf{75.8} & \textbf{92.3} & \textbf{94.5} & \textbf{65.3} & \textbf{83.6} & \textbf{82.3} & \textbf{4.9} \\
    \hline
  \end{tabular}
\end{table}

\begin{table}[htbp]
  \centering
  \scriptsize 
  \caption{Comparison of experimental performance of different methods on EL production line 3}
  \label{table 4}
  \begin{tabular}{m{3em}<{\centering} m{5em}<{\centering} m{2.5em}<{\centering} m{2.5em}<{\centering} m{2.5em}<{\centering} m{2.5em}<{\centering} m{2.5em}<{\centering} m{2.5em}<{\centering} m{2.5em}<{\centering}}
    \hline
    \multicolumn{2}{c}{Detector} & scratch & \makecell[c]{broken\\gate} & \makecell[c]{dark\\sport} & crack & \makecell[c]{unjointed\\weld} & all & GFLOPS \\
    \hline
    \multirow{3}[2]{*}{\makecell[c]{Two-stage\\detectors}} & Sparse R-CNN & 44.3 & 61.5 & 25.7 & 43.6 & 55.8 & 46.2 & 61.9 \\
          & Cascade R-CNN & 53 & 63.8 & 36.5 & 49.7 & 56.6 & 51.9 & 55.5 \\
          & SSN & 61.5 & 78.9 & 46.8 & 57.4 & 65.9 & 62.1 & 60.4 \\
    \hline
    \makecell[c]{Trans-\\former\\method} & Deformable detr & 48.3 & 51.4 & 34 & 32.9 & 47.2 & 42.7 & 65.7 \\
    \hline
    \multirow{8}[4]{*}{\makecell[c]{One-stage\\detectors}} & RetinaNet & 37 & 62.4 & 16.8 & 30.3 & 41.9 & 37.7 & 38.2 \\
          & TOOD & 51.9 & 69.7 & 29 & 53 & 61.1 & 53 & 56.8 \\
          & YOLOF & 19.1 & 55.5 & 25.8 & 27.1 & 24.1 & 30.3 & 15.1 \\
          & YOLOX & 45.9 & 62 & 36.5 & 41.7 & 45.8 & 46.4 & 14.6 \\
          & YOLOV5 & 64.8 & 91.3 & 84.3 & 80.2 & 61.2 & 76.4 & 7.1 \\
          & YOLOV7 & 65.5 & 90.7 & 83.3 & 74.9 & 53.8 & 73.6 & 10.3 \\
          & YOLOV8 & 61.4 & 89.9 & 91 & 64.4 & 68.4 & 75 & 8.2 \\
\cline{2-9}          & \textbf{Ours} & \textbf{60.1} & \textbf{90.7} & \textbf{92.3} & \textbf{80.8} & \textbf{75.8} & \textbf{79.9} & \textbf{4.9} \\
    \hline
  \end{tabular}
\end{table}

As shown in Table \ref{table 2}, Table \ref{table 3} and Table \ref{table 4}: The performance of different models trained on production line 1 exhibits varying degrees of degradation on production lines 2 and 3. This indicates the importance of enhancing the model's resilience to endogenous shift variations. Compared to the current state-of-the-art method DGGD, our method exhibits a 1\% lower detection accuracy on production line 1, but it surpasses DGGD by 9.2\% on production line 2 and 17.8\% on production line 3. This proves the superior generalization capability of our method in handling endogenous shift phenomena. Furthermore, our detection performance surpasses that of other existing methods and shows the strongest resistance to endogenous shift variations.

\subsection{Different loss functions comparison}

To validate the effectiveness of the NWD in measuring bounding box similarity for photovoltaic defects, we compared it with several commonly used bounding box loss functions. These commonly used loss functions include: Intersection over Union (IOU), Distance IOU (Diou), Generalized IOU (Giou), Exponential IOU (Eiou), Square IOU (Siou), and Complete IOU (Ciou). The experimental results are presented in Table \ref{table 5}.

\begin{table}[htbp]\scriptsize
  \centering
  \caption{Experimental results of different loss functions}
  \label{table 5}
  \setlength{\tabcolsep}{3pt}
    \begin{tabular}{m{2.5em}<{\centering}|m{2.5em}<{\centering}m{2.5em}<{\centering}m{2.5em}<{\centering}|m{2.5em}<{\centering}m{2.5em}<{\centering}m{2.5em}<{\centering}|m{2.5em}<{\centering}m{2.5em}<{\centering}m{2.5em}<{\centering}}
    \hline
    \multirow{2}[4]{*}{Loss } & \multicolumn{3}{m{8.3em}|}{EL production line 1} & \multicolumn{3}{m{8.3em}|}{EL production line 2} & \multicolumn{3}{m{8.3em}}{EL production line 3} \\
\cline{2-10}    \multicolumn{1}{c|}{} & \multicolumn{1}{m{2.5em}}{\makecell[c]{P(\%)}} & \multicolumn{1}{m{2.5em}}{\makecell[c]{R(\%)}} & \multicolumn{1}{m{2.5em}|}{\makecell[c]{mAP@\\50(\%)}} & \multicolumn{1}{m{2.5em}}{\makecell[c]{P(\%)}} & \multicolumn{1}{m{2.5em}}{\makecell[c]{R(\%)}} & \multicolumn{1}{m{2.5em}|}{\makecell[c]{mAP@\\50(\%)}} & \multicolumn{1}{m{2.5em}}{\makecell[c]{P(\%)}} & \multicolumn{1}{m{2.5em}}{\makecell[c]{R(\%)}} & \multicolumn{1}{m{2.5em}}{\makecell[c]{mAP@\\50(\%)}} \\
    \hline
    IOU   & 85.1  & 83.2  & 89.9  & 76.9  & 74.9  & 79.7  & 73.6  & 71.2  & 78 \\
    Diou  & 84.8  & 83.3  & 89.8  & 75    & 71.7  & 78    & 76.8  & 67.5  & 79.3 \\
    Giou  & 83.9  & 82.8  & 88.8  & 76.9  & 73.8  & 78.8  & 77.2  & 62.7  & 77.4 \\
    Eiou  & 84    & 83.4  & 89.6  & 79.2  & 70.4  & 78.8  & 74.5  & 63.8  & 76.7 \\
    Siou  & 84    & 83.9  & 89.8  & \multicolumn{1}{p{2.5em}}{75.9} & 76.2  & 79.3  & 66.7  & 76.7  & 77.5 \\
    Ciou  & 83.1  & 84.8  & 89.3  & 80.2  & 76.5  & 81.8  & 71.3  & 74.1  & 79.4 \\
    NWD   & 86.1  & 85.2  & 91.2  & 77.4  & 76.3  & 82.3  & 77.2  & 71.3  & 79.9 \\
    \hline
    \end{tabular}
\end{table}

From Table \ref{table 5}, it can be observed that compared to the IOU metric, the NWD metric improved by 1.3\%, 2.6\%, and 1.9\% on production lines 1, 2, and 3 respectively. Compared to the CIOU metric, the NWD metric improved by 1.9\%, 0.5\%, and 0.5\% on production lines 1, 2, and 3 respectively. Additionally, compared to other loss functions, the NWD achieved the best performance on all three production lines.

\begin{table}[htbp]\scriptsize
  \centering
  \caption{The impact of the balancing factor $\beta$ on detection performance}
    \begin{tabular}{c|ccc|ccc|ccc}
    \hline
    \multicolumn{1}{c|}{\multirow{2}[4]{*}{$\beta$}} & \multicolumn{3}{p{9.165em}|}{EL production line 1} & \multicolumn{3}{p{9.165em}|}{EL production line 2} & \multicolumn{3}{p{9.165em}}{EL production line 3} \\
\cline{2-10}          & \multicolumn{1}{p{2.055em}}{P(\%)} & \multicolumn{1}{p{2.055em}}{R(\%)} & \multicolumn{1}{p{3.055em}|}{\makecell[c]{mAP@\\50(\%)}} & \multicolumn{1}{p{2.055em}}{P(\%)} & \multicolumn{1}{p{2.055em}}{R(\%)} & \multicolumn{1}{p{3.055em}|}{\makecell[c]{mAP@\\50(\%)}} & \multicolumn{1}{p{2.055em}}{P(\%)} & \multicolumn{1}{p{2.055em}}{R(\%)} & \multicolumn{1}{p{3.055em}}{\makecell[c]{mAP@\\50(\%)}} \\
    \hline
    0.3   & 83.5  & 85.5  & 89.7  & 76.1  & 72    & 79.5  & 77.7  & 65.9  & 77.8 \\
    0.4   & 82.7  & 84.4  & 90    & 79.4  & 72.5  & 80.3  & 73.7  & 73.2  & 79.3 \\
    0.45  & 84    & 84.2  & 90.2  & 77.1  & 77.3  & 80.8  & 75.1  & 68.3  & 79.5 \\
    0.5   & 86.1  & 85.2  & \textbf{91.2} & 77.4  & 76.3  & \textbf{82.3} & 77.2  & 71.3  & 79.9 \\
    0.55  & 84.1  & 84.1  & 90.1  & 77.7  & 75.8  & 80.1  & 72.6  & 74    & \textbf{80.1} \\
    0.6   & 82.6  & 85.1  & 89.3  & 77.7  & 75.8  & 80.1  & 72.4  & 68.9  & 79.1 \\
    0.8   & 83.7  & 83.4  & 89.2  & 75.8  & 73.9  & 78.9  & 72.9  & 68.7  & 75.9 \\
    \hline
    \end{tabular}%
  \label{table 6}%
\end{table}%

\begin{figure}
	\centering
		\includegraphics[scale=0.25]{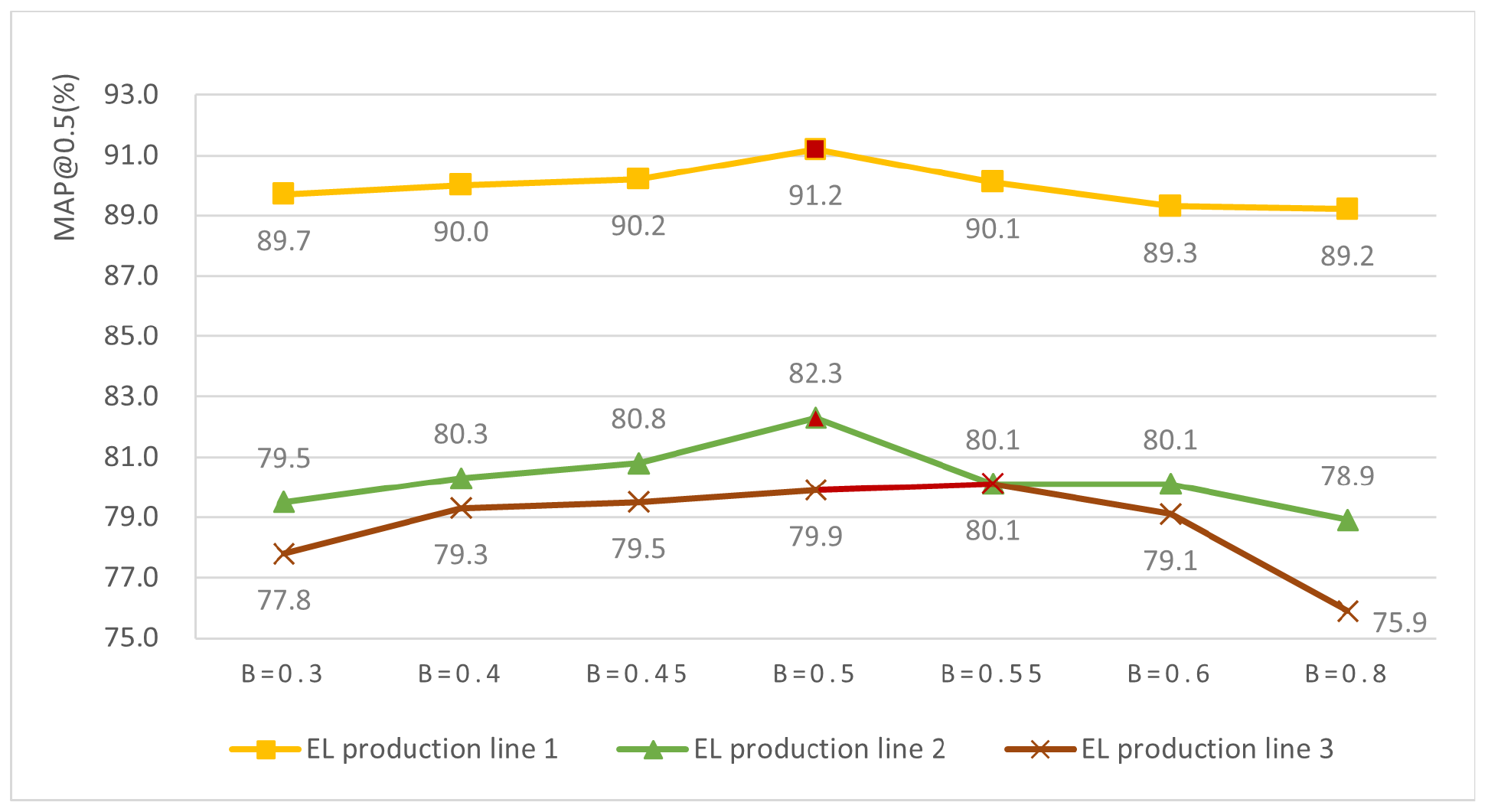}
	\caption{A schematic diagram illustrating the variation of detection performance across three production lines with different values of the balancing factor $\beta$. We have annotated the optimal values using red shading.}
	\label{Fig.9}
\end{figure}

Additionally, we also evaluated the impact of different values of the balancing factor $\beta$ on the detection performance. The effects of various values on three production lines are depicted in Table \ref{table 6} and Fig. \ref{Fig.9}.

\renewcommand{\arraystretch}{1.4} 
\begin{table}[htbp]\scriptsize
  \centering
  \caption{Display of performance details of various network modules on different domains}
    \begin{tabular}{p{3em}<{\centering}|p{6.5em}<{\centering}|p{2em}<{\centering}p{2em}<{\centering}p{2.2em}<{\centering}p{4.2em}<{\centering}p{2em}<{\centering}p{2.1em}<{\centering}}
    \hline
    \multicolumn{1}{p{1.055em}|}{\makecell[c]{Production\\line}} & \multicolumn{1}{p{1.055em}|}{\makecell[c]{Detector}} & \multicolumn{1}{p{1.055em}}{\makecell[c]{P(\%)}} & \multicolumn{1}{p{1.055em}}{\makecell[c]{R(\%)}} & \multicolumn{1}{p{1.455em}}{\makecell[c]{mAP@\\0.5(\%)}} & \multicolumn{1}{p{1.455em}}{\makecell[c]{Parameters}} & \multicolumn{1}{p{1.455em}}{\makecell[c]{GFL\\OPS}} & \multicolumn{1}{p{1.055em}}{\makecell[c]{FPS}} \\
    \hline
    \multicolumn{1}{c|}{\multirow{4}[2]{*}{\makecell[c]{EL\\production\\line 1}}} & Baseline & 82.1 & 83.7 & 88.5 & 3267590 & 8.2 & 103.7 \\
          & D2 & 82.9& 84.8 & 89.6 & 4777064 & 8.1 & 118.6 \\
          & D2+GD & 83.1 & 84.8  & 89.9  & 6646434 & 4.9 & 85.1 \\
          & D2+GD+NWD & 86.1 & 85.2 & 91.2 & 6646434 & 4.9 & 84.8 \\
    \hline
    \multicolumn{1}{c|}{\multirow{4}[2]{*}{\makecell[c]{EL\\production\\line 2}}} & Baseline &71.3& 72.3  & 76    & 3267590 & 8.2   & 103.7 \\
          & D2    &79.4& 75.7  & 80.8  & 4777064 & 8.1   & 118.6 \\
          & D2+GD &80.2& 76.5  & 81.8  & 6646434 & 4.9   & 85.1 \\
          & D2+GD+NWD &77.4& 76.3  & 82.3  & 6646434 & 4.9   & 84.8 \\
    \hline
    \multicolumn{1}{c|}{\multirow{4}[2]{*}{\makecell[c]{EL\\production\\line 3}}} & Baseline &71.4& 66.8  & 75    & 3267590 & 8.2   & 103.7 \\
          & D2    &72.8& 68.3  & 78.3  & 4777064 & 8.1   & 118.6 \\
          & D2+GD &71.3& 74.1  & 79.4  & 6646434 & 4.9   & 85.1 \\
          & D2+GD+NWD &77.2& 71.3  & 79.9  & 6646434 & 4.9   & 84.8 \\
    \hline
    \end{tabular}%
  \label{table 7}%
\end{table}%

\begin{figure}
	\centering
		\includegraphics[scale=0.3]{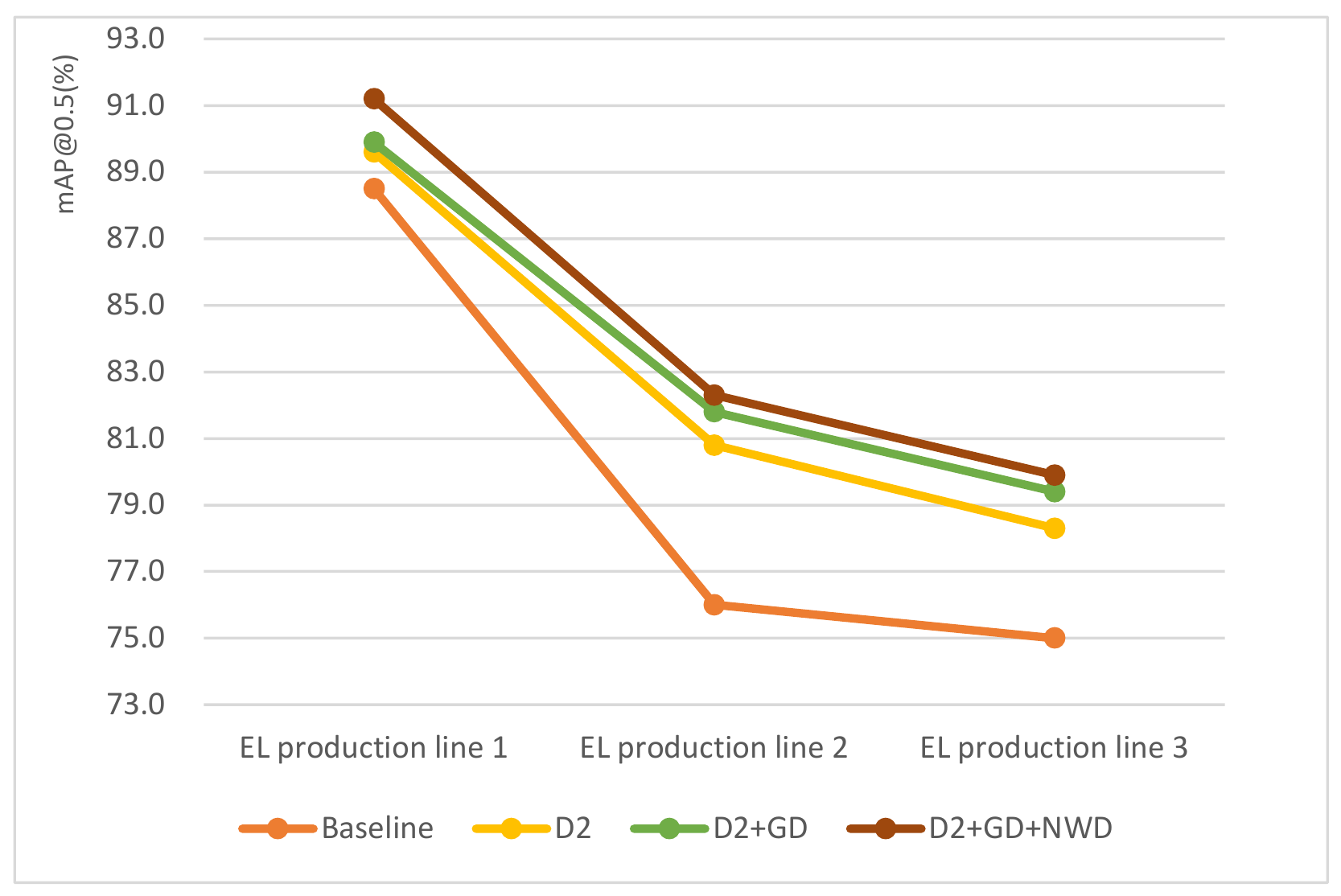}
	\caption{Comparison of performance details of various network modules in different domains.}
	\label{Fig.10}
\end{figure}

\begin{figure*}
	\centering
		\includegraphics[scale=0.32]{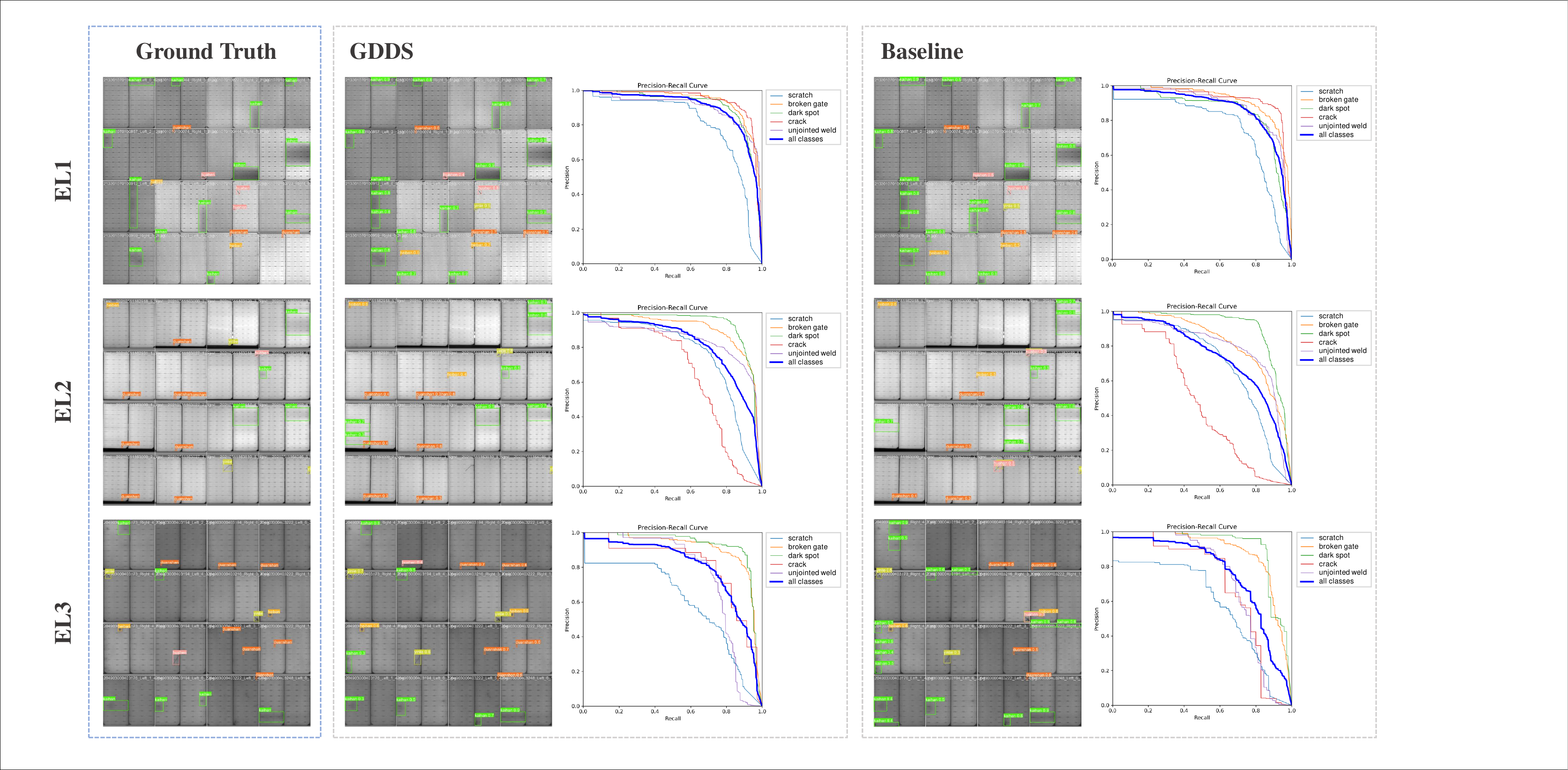}
	\caption{The visualization of the detection performance of the GDDS and baseline model on the EL Endogenous Shift Dataset. Additionally, we presented the corresponding PR curves of each network on different production lines.}
	\label{Fig.11}
\end{figure*}

\subsection{Ablation Study}

To better demonstrate the contribution of each component, we conducted an ablation analysis on the proposed model. We systematically removed or modified individual components and combinations of components and trained the models on production line 1, then evaluated their performance on production lines 2 and 3. The results of the ablation experiments are presented in Table \ref{table 7} and Fig. \ref{Fig.10}.

Compared to the baseline network, adding network hierarchy D2 resulted in an accuracy improvement of 1.1\%, 4.8\%, and 3.3\% on production lines 1, 2, and 3 respectively. Incorporating the gather-distribute mechanism on top of the added network hierarchy D2 further improved the accuracy by 0.3\%, 1\%, and 1.1\% on production lines 1, 2, and 3 respectively. This demonstrates that increasing network hierarchy and incorporating the gather-distribute mechanism enhance the model's ability to suppress unknown domain endogenous phenomena. Furthermore, adopting the improved NWD measurement method on this basis led to additional accuracy improvements of 1.3\%, 0.5\%, and 0.5\% on production lines 1, 2, and 3 respectively. This indicates that the improved measurement method is more suitable for multi-scale defect targets.

Overall, the GDDS method, compared to the baseline network, achieved accuracy improvements of 2.7\%, 6.3\%, and 4.9\% on production lines 1, 2, and 3 respectively. Additionally, the computational complexity of the model decreased by 3.3G compared to the baseline network. This fully demonstrates the superiority of the GDDS algorithm in the single-domain generalized photovoltaic defect detection task, achieving a dual enhancement in computational accuracy and speed.

\begin{figure*}
	\centering
		\includegraphics[scale=0.32]{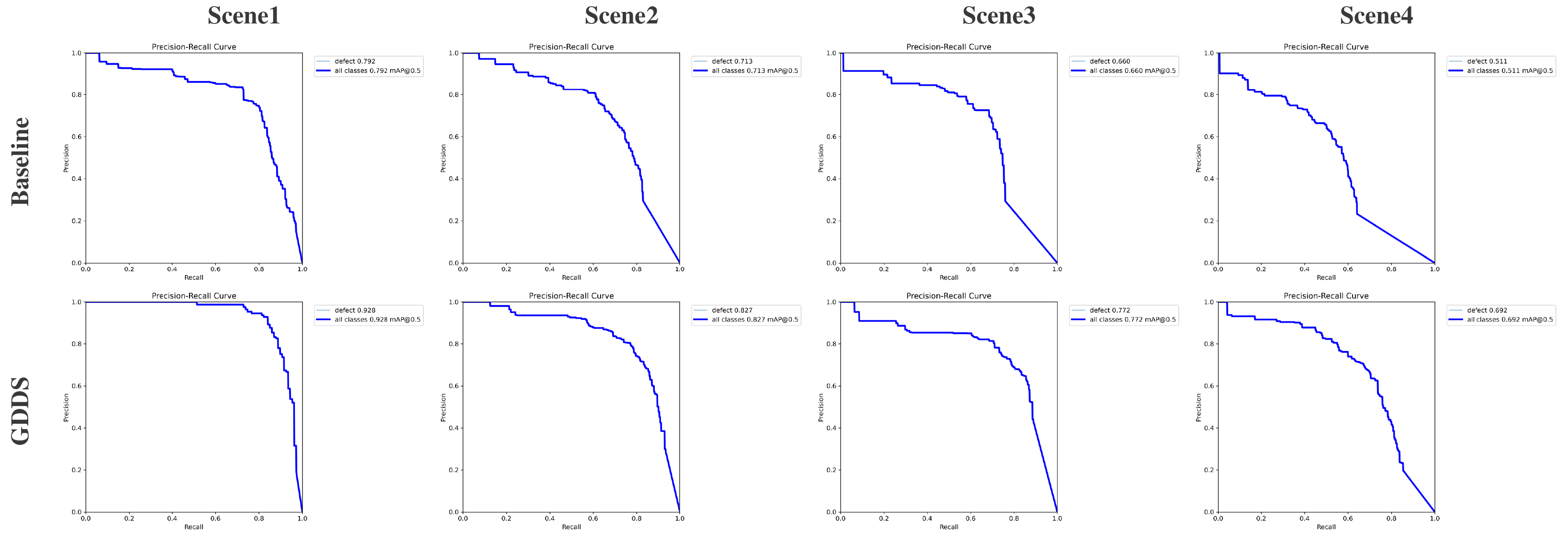}
	\caption{Visualization of the GDDS's detection performance comparison on the photovoltaic inspection infrared image dataset through PR curve graphs.}
	\label{Fig.12}
\end{figure*}

\begin{figure}
	\centering
		\includegraphics[scale=0.22]{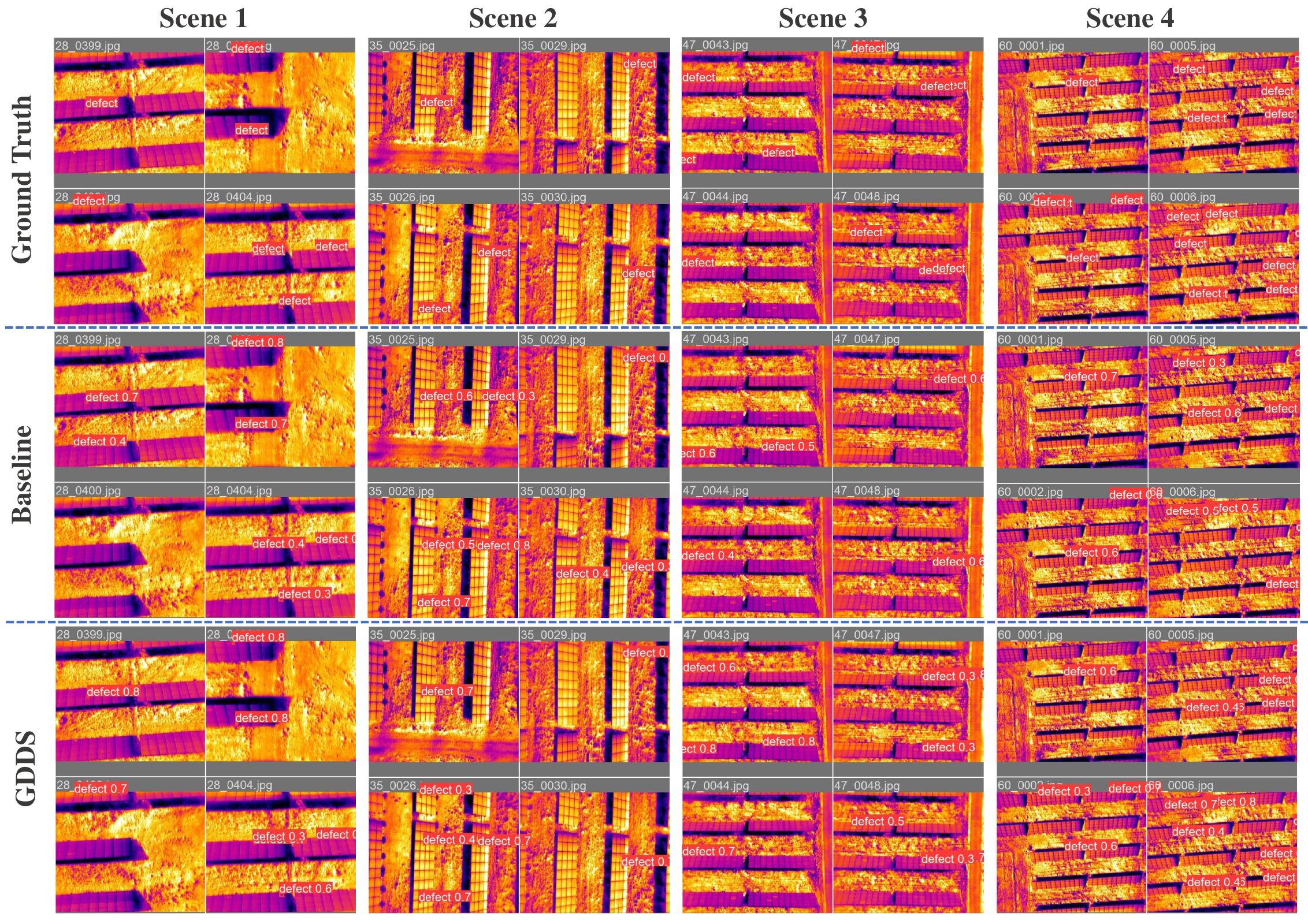}
	\caption{The visualization of the detection performance of the GDDS and baseline model on the EL Endogenous Shift DatVisualization of the detection performance of the GDDS on the photovoltaic inspection infrared image dataset.}
	\label{Fig.13}
\end{figure}

\renewcommand{\arraystretch}{1.4} 
\begin{table*}[htbp]\scriptsize
  \centering
  \caption{Comparison of experimental performance among different methods on the photovoltaic inspection infrared image dataset}
    \begin{tabular}{p{7.055em}<{\centering}p{4.5em}<{\centering}|p{2.5em}<{\centering}p{2.5em}<{\centering}p{5.0em}<{\centering}|p{2.5em}<{\centering}p{2.5em}<{\centering}p{5.0em}<{\centering}|p{2.5em}<{\centering}p{2.5em}<{\centering}p{5.0em}<{\centering}|p{2.5em}<{\centering}p{2.5em}<{\centering}p{5.0em}<{\centering}}
    \hline
    \multirow{2}[3]{*}{Detector} & \multicolumn{1}{c|}{\multirow{2}[3]{*}{GFLOPS}} & \multicolumn{3}{p{8.165em}|}{\makecell[c]{Scene 1}} & \multicolumn{3}{p{8.165em}|}{\makecell[c]{Scene 2}} & \multicolumn{3}{p{8.165em}|}{\makecell[c]{Scene 3}} & \multicolumn{3}{p{8.165em}}{\makecell[c]{Scene 4}} \\
\cline{3-14}    \multicolumn{1}{c}{} &       & \multicolumn{1}{p{2.055em}}{P(\%)} & \multicolumn{1}{p{2.055em}}{R(\%)} & \multicolumn{1}{p{4.555em}|}{mAP@50(\%)} & \multicolumn{1}{p{2.055em}}{P(\%)} & \multicolumn{1}{p{2.055em}}{R(\%)} & \multicolumn{1}{p{4.555em}|}{mAP@50(\%)} & \multicolumn{1}{p{2.055em}}{P(\%)} & \multicolumn{1}{p{2.055em}}{R(\%)} & \multicolumn{1}{p{4.555em}|}{mAP@50(\%)} & \multicolumn{1}{p{2.055em}}{P(\%)} & \multicolumn{1}{p{2.055em}}{R(\%)} & \multicolumn{1}{p{4.555em}}{mAP@50(\%)} \\
    \hline
    \makecell[c]{Faster R-CNN} & 45.4  & 73.1  & 76.3  & 75.1  & 63.1  & 59.7  & 60.7  & 62.9  & 52.1  & 52.3  & 55.8  & 51.5  & 42.2 \\
    \makecell[c]{Cascade R-CNN} & 46.8  & 73.6  & 77.7  & 76.4  & 67.9  & 63.2  & 62.5  & 67.6  & 51.7  & 54.1  & 51.5  & 46.8  & 43.9 \\
    \makecell[c]{Sparse R-CNN} & 52.1  & 63.5  & 72.2  & 63.3  & 62.2  & 59    & 59.2  & 59.2  & 47.6  & 48.7  & 54    & 34.9  & 37.5 \\
    \makecell[c]{Deformable detr} & 47.3  & 75.5  & 72.9  & 80.7  & 70    & 63.3  & 64.6  & 64.7  & 59.4  & 60.4  & 56.1  & 54.4  & 47.3 \\
    YOLOV5 & 7.1   & 71.3  & 79.1  & 77.3  & 72.8  & 71    & 70.1  & 63.5  & 72.2  & 63.3  & 63.3  & 53.1  & 45.1 \\
    YOLOV8 & 8.1   & 74.7  & 79.9  & 79.2  & 77.1  & 62.4  & 71.3  & 70.3  & 68.5  & 66    & 63.8  & 50.5  & 51.1 \\
    \textbf{Our} & \textbf{6.7} & \textbf{92.7} & \textbf{83.6} & \textbf{92.8} & \textbf{80.4} & \textbf{76.8} & \textbf{82.7} & \textbf{80.2} & \textbf{70.6} & \textbf{77.2} & \textbf{68.9} & \textbf{67.5} & \textbf{69.2} \\
    \hline
    \end{tabular}%
  \label{table 9}%

\end{table*}%

\subsection{Detection Results Visualization}

To visually demonstrate the generalization effectiveness of our model across different domains, we have visualized the detection results of models trained on different domains in Fig. \ref{Fig.11}. It can be observed that due to the complex background of photovoltaic panels, the baseline network tends to incorrectly detect many defect-free images as defective and exhibits instances of missed detections. This severely impacts the efficiency of defect detection in smart manufacturing scenarios. In contrast, our method effectively reduces instance shift caused by endogenous phenomena.

\subsection{Model Generalization Extension Experiment}

Furthermore, due to reasons such as internal battery cell damage, connection line faults, dirt, or obstructions, hotspot defects may occur during the operation of photovoltaic (PV) modules. Scanning PV modules with an infrared thermal imager and inspecting the infrared images of PV power stations can help identify potential faults. However, due to background interference factors such as weather conditions, sunlight intensity, and ambient temperature at PV power stations, defect detection in different inspection scenarios is as complex as quality inspection of PV modules in different manufacturing scenarios.

To demonstrate the generalization capability of our model across different environmental scenarios and detection tasks, we tested our proposed model and the baseline model using infrared images from inspected PV power stations in Table \ref{table 9}. Through Fig. \ref{Fig.12} and Fig. \ref{Fig.13}, we present the generalization performance of our model across multiple PV power station inspection scenarios.

From the Table \ref{table 9}, we observe a significant decrease in the detection capability of the model as the height increases. GDDS demonstrates the strongest generalization performance for detecting minor defects. In scenarios 1, 2, 3, and 4, it outperforms the baseline network by 13.6\%, 11.4\%, 11.2\%, and 18.1\%, respectively, while reducing the model's computational
complexity by 1.4G. As shown in Figure 17, with an increase in the shooting scene height, the baseline network model exhibits increasingly severe missed detection issues. The improved GDDS model better addresses the problem of missed detections, albeit with a slight risk of occasionally misclassifying extremely small photovoltaic components as defects. In practical inspection processes, defects that are extremely minor may not warrant repair.

In summary, the improved GDDS model demonstrates outstanding generalization performance in both industrial production manufacturing for photovoltaic panel quality inspection and in the practical operational processes for photovoltaic component inspection.

\section{CONCLUSION}

We propose a photovoltaic panel surface defect detection model that addresses the phenomenon of endogenous drift using a single-domain generalization approach for the first time, with a one-stage network as the baseline network. By efficiently integrating and interacting with information across layers, our model better learns the spatial features of the targets. This enables improved differentiation between targets and backgrounds in complex backgrounds, enhances adaptability to background shifts, and improves the discriminative and localization abilities for defect instance drift. Additionally, utilizing a measurement method based on NWD reduces sensitivity to position deviations of defects across multiple scales. Comprehensive experimental evaluations demonstrate that our model significantly outperforms current mainstream methods, making it more suitable for intelligent and efficient defect detection in actual photovoltaic component manufacturing.

\bibliographystyle{model1-num-names}

\bibliography{bibliography}

\end{document}